\def\mode{lncs}
\def\icpm{%
\documentclass[conference]{./tpl/icpm/IEEEtran}		
\bibliographystyle{./tpl/icpm/IEEEtran}
\usepackage[noadjust]{cite}
\renewcommand{\citepunct}{,\penalty\citepunctpenalty\,}
\renewcommand{\citedash}{--}
}
\def\lncs{
\documentclass[runningheads,orivec]{./llncs}		
\setlength{\paperheight}{232.8mm}
\setlength{\paperwidth}{151.5mm}
\setlength\voffset     {-23mm}
\setlength\hoffset     {-33mm}
\bibliographystyle{./splncs04}
}
\def\article{%
\documentclass{article}	 
\setlength{\paperheight} {232.8mm}
\setlength{\paperwidth}  {151.5mm}
\setlength\voffset       {-23mm}
\setlength\hoffset       {-34mm}
\bibliographystyle{plain}
\usepackage[firstpage]{draftwatermark}
\SetWatermarkAngle{0}
\SetWatermarkFontSize{10pt}
\SetWatermarkHorCenter{153mm}
\SetWatermarkVerCenter{247mm}
\SetWatermarkLightness{0.6}
\SetWatermarkText{Draft{,} {\monthyeardate\today}}
}
\csname\mode\endcsname

\newif\ifshowtodos
\showtodostrue		

\newcommand{\authorartem}		 {Artem~Polyvyanyy}
\newcommand{\authoralistair} {Alistair~Moffat}
\newcommand{\authorluciano}	 {Luciano~Garc{\'{\i}}a{-}Ba{\~{n}}uelos}

\newcommand{\articletitle}   {Bootstrapping Generalization of\\ Process Models Discovered From Event Data}
\newcommand{\articlesubjet}  {Computer Science, Process Mining, Process Querying, Information Systems}
\newcommand{\articleauthors} {\authorartem, \authoralistair, \authorluciano}

\usepackage[usenames,dvipsnames]{xcolor}
\usepackage[algoruled,vlined,linesnumbered]{algorithm2e}
\usepackage{amsmath}
\usepackage{amsfonts}
\usepackage{mdframed}
\usepackage{paralist}
\setdefaultleftmargin{1em}{1em}{1em}{1em}{1em}{1em}
\usepackage{datetime}
\newdateformat{monthyeardate}{\monthname[\THEMONTH] \THEYEAR}
\usepackage[pdfsubject={\articlesubjet},
                            pdftitle={\articletitle},
                            pdfauthor={\articleauthors},
                            pdfkeywords={\articlesubjet},
                            colorlinks=false,
                            pdfborder={0.25 0.25 0.25}]{hyperref}
\hypersetup{bookmarksdepth}
\usepackage[algoruled]{algorithm2e}
\ifshowtodos
\usepackage[textsize=scriptsize,colorinlistoftodos]{todonotes}
\else
\usepackage[textsize=scriptsize,colorinlistoftodos,disable]{todonotes} 
\fi
\let\todonote\todo
\renewcommand{\todo}[2]{\todonote[inline,color=red!20]{TODO (#1): #2}}

\usepackage{tikz,pgf,xcolor}
\usetikzlibrary{arrows,automata,trees,plotmarks,shadows,shapes}
\usepackage{pgfplots}
\usetikzlibrary{pgfplots.groupplots}
\usepackage{url}
\usepackage{mathtools} 
\usepackage[capitalise,nameinlink]{cleveref}
\usepackage{wrapfig}

\definecolor{mybluecolor}{RGB}{50,106,218}
\definecolor{myredcolor}{RGB}{176,53,53}
\definecolor{mygreencolor}{RGB}{93,172,0}
\definecolor{myyellowcolor}{RGB}{255,163,34}
\definecolor{mypurplecolor}{RGB}{86,35,132}
\definecolor{mytealcolor}{RGB}{30,161,165}

\newcommand{\splitatcommas}[1]{%
  \begingroup
  \ifnum\mathcode`,="8000
  \else
    \begingroup\lccode`~=`, \lowercase{\endgroup
      \edef~{\mathchar\the\mathcode`, \penalty0 \noexpand\hspace{-1pt plus 3em}}%
    }\mathcode`,="8000
  \fi
  #1%
  \endgroup
}

\newcommand{\func}[3]{{{#1}:{#2} \rightarrow {#3}}}
\newcommand{\funcCall}[2]{{\ensuremath {\mathit{#1}}_{\!}\left({#2}\right)}}

\DeclarePairedDelimiter\ceil{\lceil}{\rceil}

\newcommand{\intintervalcc}[2]{{\ensuremath \left[#1 \,..\, #2\right]}}
\newcommand{\intervalcc}[2]{{\ensuremath \left[#1, #2\right]}}

\providecommand{\powerset}[1]{{\ensuremath \mathcal{P}\!\left({#1}\right)}}
\renewcommand{\powerset}[1]{{\ensuremath \mathcal{P}\!\left({#1}\right)}}

\providecommand{\support}[1]{{\ensuremath \mathit{Supp}\left({#1}\right)}}
\renewcommand{\support}[1]{{\ensuremath \mathit{Supp}\left({#1}\right)}}
\providecommand{\cardinality}[1]{\ensuremath \left|{#1}\right|}
\renewcommand{\cardinality}[1]{\ensuremath \left|{#1}\right|}

 \newcommand{\mset}[1] {\ensuremath [\splitatcommas{#1}]}
\newcommand{\msetel}[2]{{\ensuremath {{#1}^{#2}}}}

\newcommand{\set}[1]{\ensuremath \left\{\splitatcommas{#1}\right\}}

\newcommand{\pair}[2]{\ensuremath \left(\splitatcommas{#1, #2}\right)}
\newcommand{\triple}[3]{\ensuremath \left(\splitatcommas{#1, #2, #3}\right)}
\newcommand{\tuple}[1]{\ensuremath \left(\splitatcommas{#1}\right)}

\newcommand{\kleenestar}[1]{{\ensuremath {#1}^{*}}}
\newcommand{\emptysequence}{{\ensuremath \epsilon}}

\newcommand{\seqLength}[1]{\ensuremath \left|{#1}\right|}
\newcommand{\concat}[2]{\ensuremath #1 \circ #2}
\newcommand{\subseq}[3]{\ensuremath \funcCall{subseq}{#1, #2, #3}}
\newcommand{\subseqid}[3]{\ensuremath \triple{#1}{#2}{#3}}
\newcommand{\crossover}[3]{\ensuremath #1 \,\otimes\, #2 \,=\, #3}
\newcommand{\crossoverop}[2]{\ensuremath #1 \otimes #2}

\providecommand{\implies}{\ensuremath \Rightarrow}
\renewcommand	 {\implies}{\ensuremath \Rightarrow}

\newcommand{\action}[1]{\texttt{#1}}
\newcommand{\trace}[1]{\texttt{#1}}

\newcommand{\fig}[9]{\begin{figure}[#1]
\vspace{#2mm}
\begin{center}
	\includegraphics[scale=#3,trim=#4]{#5}
\end{center}
\vspace{#6mm}
\caption{#7.}
\vspace{#8mm}
\label{#9}
\end{figure}}

\newcommand{\orcidartem}		{\href{https://orcid.org/0000-0002-7672-1643}{\protect\includegraphics[scale=0.05]{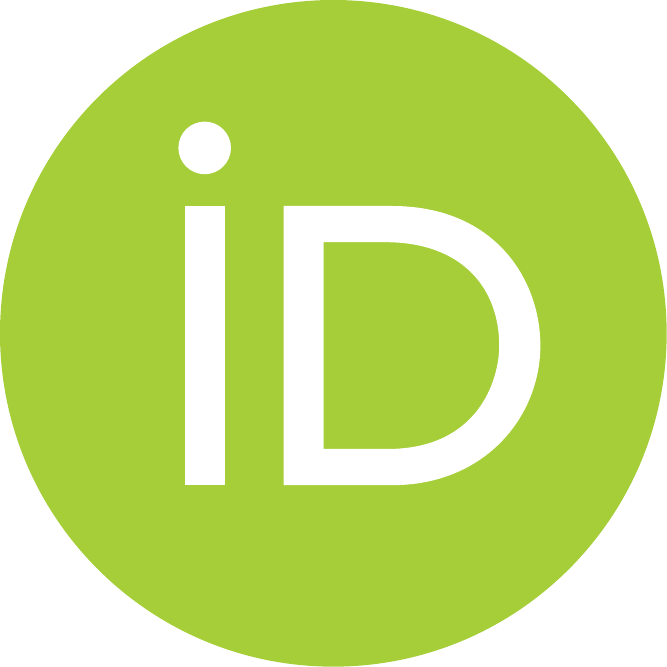}}}	
\newcommand{\orcidalistair}	{\href{https://orcid.org/0000-0002-6638-0232}{\protect\includegraphics[scale=0.05]{fig/orcid}}}	
\newcommand{\orcidluciano}	{\href{https://orcid.org/0000-0001-9076-903X}{\protect\includegraphics[scale=0.05]{fig/orcid}}}	

\newcommand{\actions}			{{\ensuremath \Lambda}}

\newcommand{\natnum}					{\ensuremath \mathbb{N}}
\newcommand{\natnumwithzero}	{\ensuremath \mathbb{N}_0}

\newcommand{\ie}					{i.e.,~}
\newcommand{\eg}					{e.g.,~}

\newcommand{\eqntopspace}{\vspace*{-1.0ex}}
\newcommand{\eqnbotspace}{\vspace*{-0.8ex}}

\newtheorem{mytheorem}		{Theorem}
\newtheorem{mydefinition}	{Definition}
\newtheorem{mylemma}			{Lemma}
\newtheorem{myproposition}{Proposition}
\newtheorem{mycorollary}	{Corollary}
\newtheorem{myexample}		{Example}
\newtheorem{myconjecture}	{Conjecture}

\newtheorem{myinvariant}	{Invariant}

\numberwithin{mytheorem}		{section}
\numberwithin{mydefinition}	{section}
\numberwithin{mylemma}			{section}
\numberwithin{myproposition}{section}
\numberwithin{mycorollary}	{section}
\numberwithin{myexample}		{section}
\numberwithin{myconjecture}	{section}
\numberwithin{myremark}			{section}
\numberwithin{myinvariant}	{section}

\newenvironment{define}[3][]
{\begin{mydefinition}[#2]\label{#3}#1\normalfont}
{\hfill\ensuremath{\lrcorner}\end{mydefinition}}

\newenvironment{lem}[3][]
{\begin{mylemma}[#2]\label{#3}#1}
{\hfill\ensuremath{\lrcorner}\end{mylemma}}

\newenvironment{thm}[3][]
{\begin{mytheorem}[#2]\label{#3}#1}
{\hfill\ensuremath{\lrcorner}\end{mytheorem}}

\addtocounter{mytheorem}{0}

\usepackage{subfig}
\usepackage{nicefrac}
\usepackage{multirow}
\usepackage{booktabs}
\usepackage{siunitx}
\usepackage{txfonts} 
\usepackage{lipsum}
\usepackage{adjustbox}

\newcommand{\updated}[1]{\textcolor{black}{#1}}

\title{\articletitle}

\date{\today}

\begin{document}

\author{
\authorartem\inst{1}~\orcidartem
\and 
\authoralistair\inst{1}~\orcidalistair
\and
\authorluciano\inst{2}~\orcidluciano
}
\titlerunning{Bootstrapping Generalization of Discovered Process Models}
\authorrunning{\authorartem, \authoralistair, and \authorluciano}

\institute{
The University of Melbourne, Victoria 3010, Australia\\
\href{mailto:artem.polyvyanyy@unimelb.edu.au;ammoffat@unimelb.edu.au}{\{artem.polyvyanyy;ammoffat\}@unimelb.edu.au}
\and
Tecnol{\'{o}}gico de Monterrey, 64849 Monterrey, N.L., Mexico\\
\href{mailto:luciano.garcia@tec.mx}{luciano.garcia@tec.mx}
}

\maketitle


\setcounter{footnote}{0}

\vspace{-4mm}

\begin{abstract}
Process mining extracts value from the traces recorded in the event logs of IT-systems, with {\emph{process discovery}} the task of inferring a process model for a log emitted by some unknown system.
{\emph{Generalization}} is one of the quality criteria applied to process models to quantify how well the model describes future executions of the system.
\updated{Generalization is also perhaps the least understood of those criteria}, with that lack primarily a consequence of it measuring properties over the entire future behavior of the system when the only available sample of behavior is that provided by the log.
In this paper, we apply a bootstrap approach from computational statistics, allowing us to define an estimator of the model's generalization based on the log it was discovered from.
We show that standard process mining assumptions lead to a {\emph{consistent estimator}} that makes fewer errors as the quality of the log increases. 
Experiments confirm the ability of the approach to support industry-scale data-driven systems engineering.

\end{abstract}

\vspace{-8mm}
\section{Introduction}
\label{sec:intro}
\vspace{-2mm}
\enlargethispage{\baselineskip}

Given an event log that records traces of some real-world system, the
challenge of {\emph{process discovery}} is to develop a plausible
{\emph{model}} of that system, so that the behavior of the system can
be analyzed independently of the specific transactions included in
that particular log.
Many different models might be constructed from the same log.
Thus, it is important to have tools that allow the quality of a given
model to be {\emph{quantified}} relative to the initial log.
For example, {\emph{precision}} is the fraction of the traces
permitted by the model that appear in the log, and {\emph{recall}} is
the fraction of the log's traces that are valid according to the
model.
Composite measures have also been defined
{\cite{Aalst2016,Carmona2018}}.

A log is only a sample of observations in regard to the underlying
system, and not a specification of its actions.
It is thus interesting to consider {\emph{generalization}} -- the
extent to which the inferred model accounts for future observations
of the system.
Generalization poses substantial challenges, since, by its very
definition, it asks about behaviors that have {\emph{not}} been
observed from a system that is {\emph{not}} known.
High generalization (and high recall) can be obtained by allowing all
possible traces.
But overly-permissive models of necessity compromise precision.
What is desired is a model that attains high precision and recall
with respect to the supplied log, and continues to score well on the
universe of possible logs that might arise via continued
observation.
{\updated{Note that process mining generalization as studied in this
work differs from generalization as it applies to process model
abstraction~\cite{PolyvyanyySW08}.
Process model abstraction considers techniques for combining
several processes, activities, and events into corresponding
generalized concepts, for example, identifying a semantically
coherent sub-process in a process model.}}

\updated{In particular}, we study \updated{the problem of measuring
the generalization of a discovered process model, making use of the
{\emph{bootstrapping}} technique from computational
statistics~{\cite{Efron1993}}.}
In the simplest form, the idea is to construct multiple
sampled replicates of the initial log, each representing a log that
might have emerged from the system.
Any aggregate properties established by considering the set of
replicates can then be assumed to be valid for the universe of
possible traces.
That is, by constructing a process model from one replicate, and then
testing on another, generalization can be explored.
In terms of high-level contributions, our work here: 

\smallskip
\begin{compactitem}
\item 
Presents, for the first time, an estimator of the generalization of a
process model discovered from an event log, grounded in the bootstrap
method;
\item 
Shows that the estimator is consistent for the class of systems
captured as directly-follows graphs (DFGs), making fewer errors on
larger log replicates; and
\item
Confirms via experiments the feasibility of the new approach in
industrial settings.
\end{compactitem}
\smallskip

\noindent
The next section introduces several key ideas, and a running example.
{\cref{sec:approach}} presents our new approach, and demonstrates its
consistency. {\cref{sec:evaluation}} provides an evaluation that
confirms the consistency and feasibility of our approach.
Related work is discussed in {\cref{sec:related}}. Finally,
{\cref{sec:conclusion}} concludes our presentation.


\vspace{-2mm}
\section{Background}
\label{sec:background}
\vspace{-1mm}
\enlargethispage{\baselineskip}

\vspace{-1mm}
\subsection{Systems, Models, Logs, and Their Languages}
\label{subsec:languages}
\vspace{-1mm}

\updated{%
For the purpose of formalizing the problem of measuring
generalization of a {\emph{process model}} discovered from an
{\emph{event log}} of a {\emph{system}}, consistent with the standard
formalization in process mining {\cite{Buijs2014,Aalst2016}},
we interpret the system, model, and log as collections of traces},
where a {\emph{trace}} is a sequence of actions that attains, or
might attain, some goal.

Let $\actions$ be a set of possible {\emph{actions}}; $\actions =
\set{\action{a},\action{b},\action{c},\action{d},\action{e},\action{f}}$ will be
used throughout this section.
Define $\kleenestar{\actions}$ to be the set of all possible
{\emph{traces}} over $\actions$, each a finite sequence of
actions.
Both
$\trace{abbcf}$ and $\trace{addef}$ are traces over $\actions$, as
is the empty trace, denoted by~$\emptysequence$.

\smallskip
\noindent
\textbf{Systems.}
A {\emph{system}} $S$ is a group of active elements, such as software
components \updated{and agents},
that perform actions and thereby consume, produce, or manipulate
objects \updated{and} information.
\updated{%
A system can be an information system or a business process with its
organization context, business rules, and
resources~\cite{Buijs2014}.}
Any sequence of actions that leads to the system's goal constitutes a
trace.
In general, a system might generate an infinite collection of traces,
possibly containing infinitely many distinct traces, and hence also
possibly containing traces of arbitrary length.

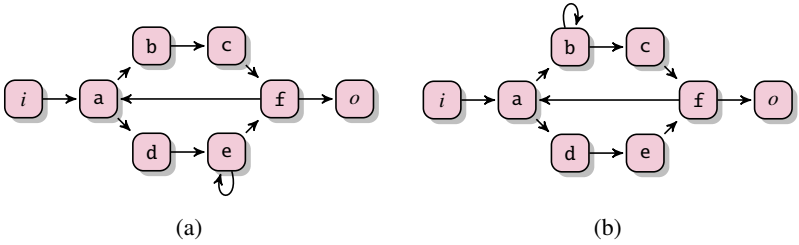
\begin{figure*}[t]
\centering
\subfloat[]{
\begin{tikzpicture}[scale=0.55, transform shape, ->, >=stealth', shorten >=1pt, auto, node distance=18mm, on grid, semithick, action/.style={fill=purple!20, draw, rounded corners, minimum size=9mm, drop shadow}]
\node[action]	(n0) 											{\Large $i$};
\node[action]	(n1) [right=of n0] 				{\Large $\texttt{a}$};
\node[action] (n2) [above right=of n1]	{\Large $\texttt{b}$};
\node[action] (n3) [below right=of n1]	{\Large $\texttt{d}$};
\node[action] (n4) [right=of n2]				{\Large $\texttt{c}$};
\node[action] (n5) [right=of n3]				{\Large $\texttt{e}$};
\node[action] (n6) [below right=of n4]	{\Large $\texttt{f}$};
\node[action] (n7) [right=of n6]				{\Large $o$};

\path (n0) edge node {\small} (n1)
			(n1) edge node {\small} (n2)
					 edge node {\small} (n3)
			(n2) edge node {\small} (n4)
			(n3) edge node {\small} (n5)
			(n4) edge node {\small} (n6)
			(n5) edge [loop below] node {\small} (n5)
					 edge node {\small} (n6)		 
			(n6) edge node {\small} (n7)		 
			(n6) edge node {\small} (n1)
;
\end{tikzpicture}
\label{fig:example:process:model}%
}
\hspace{4mm}
\subfloat[]{
\begin{adjustbox}{trim=0 0 0 2.75cm}
\begin{tikzpicture}[scale=0.55, transform shape, ->, >=stealth', shorten >=1pt, auto, node distance=18mm, on grid, semithick, action/.style={fill=purple!20, draw, rounded corners, minimum size=9mm, drop shadow}]
\node[action]	(n0) 											{\Large $i$};
\node[action]	(n1) [right=of n0] 				{\Large $\texttt{a}$};
\node[action] (n2) [above right=of n1]	{\Large $\texttt{b}$};
\node[action] (n3) [below right=of n1]	{\Large $\texttt{d}$};
\node[action] (n4) [right=of n2]				{\Large $\texttt{c}$};
\node[action] (n5) [right=of n3]				{\Large $\texttt{e}$};
\node[action] (n6) [below right=of n4]	{\Large $\texttt{f}$};
\node[action] (n7) [right=of n6]				{\Large $o$};

\path (n0) edge node {\small} (n1)
			(n1) edge node {\small} (n2)
					 edge node {\small} (n3)
			(n2) edge [loop above] node {\small} (n2)
					 edge node {\small} (n4)
			(n3) edge node {\small} (n5)
			(n4) edge node {\small} (n6)
			(n5) edge node {\small} (n6)		 
			(n6) edge node {\small} (n7)		 
			(n6) edge node {\small} (n1)
;
\end{tikzpicture}
\end{adjustbox}
\label{fig:example:system}%
}
\vspace{-3mm}
\caption{\small (a) An example process model, and (b) an example system.}
\label{fig:1}%
\vspace{-5mm}
\end{figure*}

\smallskip
\noindent
\textbf{Models.}
A {\emph{process model}}, or just a {\emph{model}}, $M$ is a finite
description of a
set of traces.
{\Cref{fig:example:process:model}} describes a process
model, represented as a {\emph{directly-follows graph}} (DFG), with
start node $i$, final node $o$, and all \updated{walks} from $i$ to $o$ as valid
traces.
That example model, for instance, describes traces $\trace{abcf}$ and
$\trace{adeef}$; but does not describe $\trace{abbcf}$.

\smallskip
\noindent
\textbf{Logs.}
An {\emph{event log}}, or just a {\emph{log}}, $L$ is a finite multiset
of traces.

\smallskip
\noindent
\textbf{Languages.}
A {\emph{language}} is a subset of the traces in
$\kleenestar{\Lambda}$.
The language of system $S$ is the set of all traces
$S$ can generate; the language of model $M$ is the set of traces
described by $M$; and the language of log $L$ is its support set,
$\support{L}$.
Further, define $\mathcal{L} \subset
\powerset{\kleenestar{\Lambda}}$, $\mathcal{M} \subseteq
\powerset{\kleenestar{\Lambda}}$, and $\mathcal{S} \subseteq
\powerset{\kleenestar{\Lambda}}$ to be sets containing all possible
languages of logs, models, and systems, respectively, with
$\mathcal{L}$ restricted to finite languages.
When the context is clear, we will interpret logs, models, and systems
as their languages -- if we say that model $M$ was discovered from
log $L$ out of system $S$, we may be referring to the concrete
system, model, and log, or may be referring to the languages they
describe.

\vspace{-2mm}
\subsection{Process Discovery}
\label{subsec:process:discovery}
\vspace{-1mm}

Given a log, the {\emph{process discovery problem}} consists
of constructing a model that represents the behavior recorded in the
log~\cite{Aalst2016}.
For example, using superscripts to indicate multiplicity,
let $L=\mset{
\msetel{\trace{abbbcf}}{5},
\msetel{\trace{abcf}}{20},
\msetel{\trace{addef}}{},
\msetel{\trace{adeef}}{10},
\msetel{\trace{adefabcfadef}}{10},
\msetel{\trace{adef}}{20}
}$
be an event log that contains six distinct traces and 66 traces in
total.
Many comprehensive process discovery techniques have been devised over the last two decades~\cite{Aalst2016}.
However, model $M$, shown in {\cref{fig:example:process:model}},
can be constructed from $L$ via a simple four-stage discovery
algorithm: (1) filter out infrequent traces by, for example, removing
the least frequent third of the distinct traces; (2) for every action
in each remaining trace, construct a node representing that action;
(3) for every pair of adjacent actions $x$ and $y$, introduce a
directed edge from the node for $x$ to the node for $y$; and (4)
introduce start node $i$ and end node~$o$, together with edges from
$i$ to every initial action in a frequent trace, and from every last
action in a frequent trace to the sink node~$o$.

Despite the simplicity of that supposed construction process, $M$
{\emph{fits}} $60$ of the traces in $L$, failing on only six.
On the other hand, the cycles in $M$ mean that it represents
infinitely many traces {\emph{not}} present in $L$.
To quantify the extent of the mismatch between $L$ and $M$, the
measures {\emph{recall}} and {\emph{precision}} can be
used~\cite{Buijs2014,Carmona2018,Polyvyanyy2020b}.
Given a suite of possible models, precision and recall allow
alternative models to be numerically compared.

\vspace{-2mm}
\subsection{Generalization}
\label{subsec:generalization}
\vspace{-1mm}
\enlargethispage{\baselineskip}

\begin{wrapfigure}[15]{R}{0.386\textwidth}
\centering
\vspace*{-0.0ex}
\includegraphics[scale=0.55]{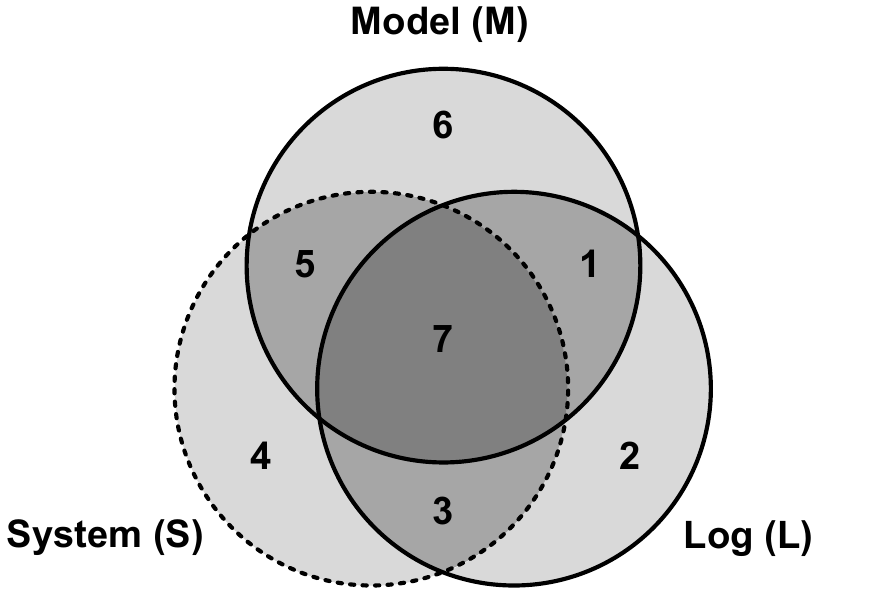}
\caption{\small Venn diagram showing languages of model $M$, log $L$, and
system $S$, adapted from Buijs et al.~\cite{Buijs2014};
the language of the system is unknown (the dotted border).
\label{fig:venn}}
\end{wrapfigure}

An event log of a system contains traces that the system generated
over some finite period {\emph{and}} were recorded using some logging
mechanism.
That is, a log is a {\emph{sample}} of all possible traces the system
could have generated~\cite{DBLP:journals/corr/abs-2012-12764}.
Hence, an alternative (and arguably more useful) definition of the
process discovery problem is that a model be constructed to represent
{\emph{all}} of the traces the system {\emph{could}} have generated,
derived from the finite sample provided in the log.
Such a model, if constructed, would explain the {\emph{system}}, and
not just the traces that happened to be recorded in that particular
log.
For example, the DFG $S$ in {\cref{fig:example:system}} could be a
complete representation of the system that generated the 66 traces
contained in $L$, allowing, for example, the five occurrences of
$\trace{abbbcf}$ to now be understood.

If the alternative definition of the process discovery problem is
accepted, then the candidate model $M$ in
{\cref{fig:example:process:model}} must be somehow benchmarked
against the system $S$ of {\cref{fig:example:system}}, rather than
against $L$.
Unfortunately, the actual behavior of the system is often unknown;
indeed, that absence is, of course, a primary motivation for process
discovery.
That is, the log may be the only available information in respect of
the system whose behavior it is a sample of.
Given this context, {\Cref{fig:venn}} shows the relationship between
the languages of log, model, and system.
The numbered regions then have the following
interpretations (again, making use of example log $L$, model $M$, and
system $S$): 
(1)
Traces that $S$ does not generate, yet appear in $L$
(perhaps by error) and are included
in $M$; \eg {\trace{adeef}}.
(2)
Traces that $S$ does not generate, yet appear in $L$
without triggering inclusion in $M$; \eg {\trace{addef}}.
(3)
Traces permitted by $S$, and recorded in $L$,
but not included in $M$; \eg
{\trace{abbbcf}}.
(4)
Traces permitted by $S$, but neither observed in
$L$ nor permitted by $M$; \eg {\trace{abbcf}}.
(5)
Traces permitted by both $S$ and $M$, but not
appearing in $L$; \eg {\trace{adefadef}}.
(6)
Traces neither permitted by $S$ nor observed in $L$,
but nevertheless allowed by $M$; \eg
{\trace{adeeef}}.
(7)
Traces permitted by $S$, observed in $L$, and included in
$M$; \eg
{\trace{adefabcfadef}}.
Note that categories (4), (5), and (6) might be infinite, but that
(1), (2), (3), and (7) must be finite, \updated{as} $L$ itself is
finite.

To assess a process model against a system, a
{\emph{generalization}} measure is employed~\cite{Aalst2016}.
The objective of a generalization measure is described
by van der Aalst~\cite{Aalst2018} as:

\begin{quote}
\vspace*{-1ex}
\it
a generalization measure [{\dots}]
aims to quantify the likelihood that new unseen [traces generated
by the system] will fit the model.
\end{quote}

On the assumption that $M \subseteq S$, Buijs et al.~\cite{Buijs2014}
suggest measuring the generalization of $M$ with respect to $S$ as
the model-system recall, that is, the fraction of the system covered
by the model, $(S \cap M)/M$ (when the context is clear, we use $X$
to denote $|X|$).
But this proposal requires knowledge of, or a way to approximate, the
system's traces.

More broadly, generalization is probably the least understood quality
criterion for discovered models in process mining.
Only a few approaches have been described, and all of them diverge,
in one way or another, from the intended
phenomenon~\cite{Syring2019}.
We elaborate on that observation in {\cref{sec:related}}, which
discusses related work.


\vspace{-3mm}
\section{Estimating Generalization}
\label{sec:approach}
\vspace{-2mm}
\enlargethispage{\baselineskip}

We now present our proposal.
{\Cref{subsec:bootstrapping}} summarizes the bootstrap method from
statistics, a key component; and {\cref{subsec:framework}} presents a
framework for measuring generalization using it.
Then, {\cref{subsec:log:sampling}} develops the required log sampling
mechanism; {\cref{subsec:measures}} presents concrete instantiations
of the framework; and {\cref{subsec:consistency}} establishes the consistency of the presented estimator. 
Finally, {\cref{subsec:example}} demonstrates
the application of our approach to the running example of
{\cref{subsec:process:discovery}}.

\vspace{-2mm}
\subsection{Bootstrapping}
\label{subsec:bootstrapping}
\vspace{-1mm}

Bootstrapping is a computational method in statistics that estimates
the sampling distribution over unknown data using the sampling
distribution over an approximate sufficient statistic of the
data~\cite{Efron1993}.
The true sampling distribution of some quantity for a population is
constructed by drawing multiple samples from the true population,
computing the quantity for each sample, and then aggregating the
quantities.
But if the true population is unknown, drawing samples may be
expensive, or even infeasible.
Instead, the bootstrap method can be used, shown in
{\cref{fig:bootstrapping}}, with dashed and solid lines denoting
unknown and observed quantities, respectively.
The bootstrap proceeds in four steps:
\smallskip
\begin{compactenum}[1.]
\item \emph{Take a single sample} of the true population.
\item \emph{Estimate the population} based on that single sample.
\item \emph{Compute many samples} from that estimated population.
\item \emph{Estimate the sampling distribution} based on those samples.
\end{compactenum}
\smallskip
Estimated sampling distributions can be used to approximate
properties of the sampled distribution, including the mean and its
confidence interval, and variance~\cite{Efron1982,Breiman1996}.

\fig{t}{-2}{0.55}{0 0 0 0}{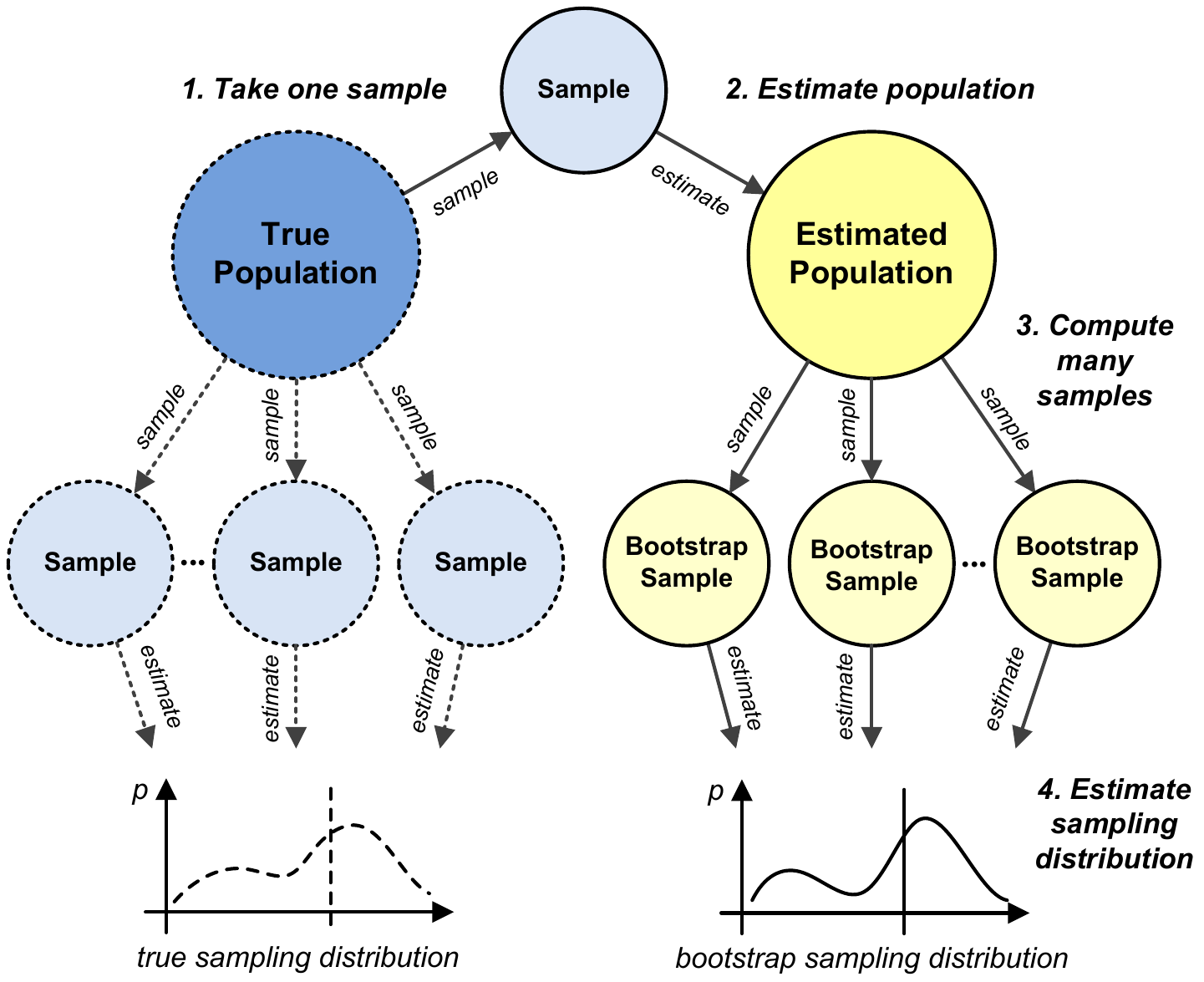}{-4}{\small The bootstrap method,
adapted from lecture notes
at the Pennsylvania State University, see
{\url{https://online.stat.psu.edu/stat555/node/119/}}, accessed 26
November 2021}{-5}{fig:bootstrapping}

\vspace{-2mm}
\subsection{Bootstrap Framework for Measuring Generalization}
\label{subsec:framework}
\vspace{-1mm}
\enlargethispage{\baselineskip}

We now apply the bootstrap method to estimate generalization of
candidate models for representing some system, supposing that for
every model the corresponding system is known, and seeking measures
of the form
$\func{gen}{\mathcal{M} \times \mathcal{S}}{\intervalcc{0}{1}}$.
The better $M$ represents $S$, the higher is $\funcCall{gen}{M,S}$;
with $\funcCall{gen}{M,S}=1$ arising if every new trace from $S$ is
described by $M$.
Conversely, $\funcCall{gen}{M,S}=0$ is the worst possible
generalization, arising when none of the new distinct traces observed
from $S$ are captured by $M$.

As the system is not known, it cannot be measured directly.
We thus propose assessing log-based generalization via an estimator
function:
\eqntopspace
\begin{equation}
\label{eq:generalization:estimator:function}
\func{gen^*}{\mathcal{M} \times \mathcal{L} \times \mathbb{LSM} \times \mathbb{N} \times \mathbb{N}}{\intervalcc{0}{1}} \,,
\eqnbotspace
\end{equation}
where $\mathbb{LSM}$ is a collection of log sampling methods, with
each $\mathit{lsm} \in \mathbb{LSM}$ a {\emph{randomizing}} function
that, given an event log $L$ and an integer $n$, produces a sample
log $L^*=\funcCall{lsm}{L,n}$ from $L$ of size $n$.
Given a model $M$, a log $L$, a log sampling method $\mathit{lsm}$, a
log sample size $n$, and a log sample count $m$,
{\cref{alg:bootstrap:gen}} implements the estimator:
\eqntopspace
\begin{equation}
\label{eq:generalization:estimator:function:definition}
\funcCall{gen^*}{M,L,\mathit{lsm},n,m}
=\funcCall{\text{BootstrapGeneralization}}{M,L,\mathit{gen},\mathit{lsm},n,m}
\,.
\eqnbotspace
\end{equation}

\fig{t}{-1}{0.55}{0 0 0 0}{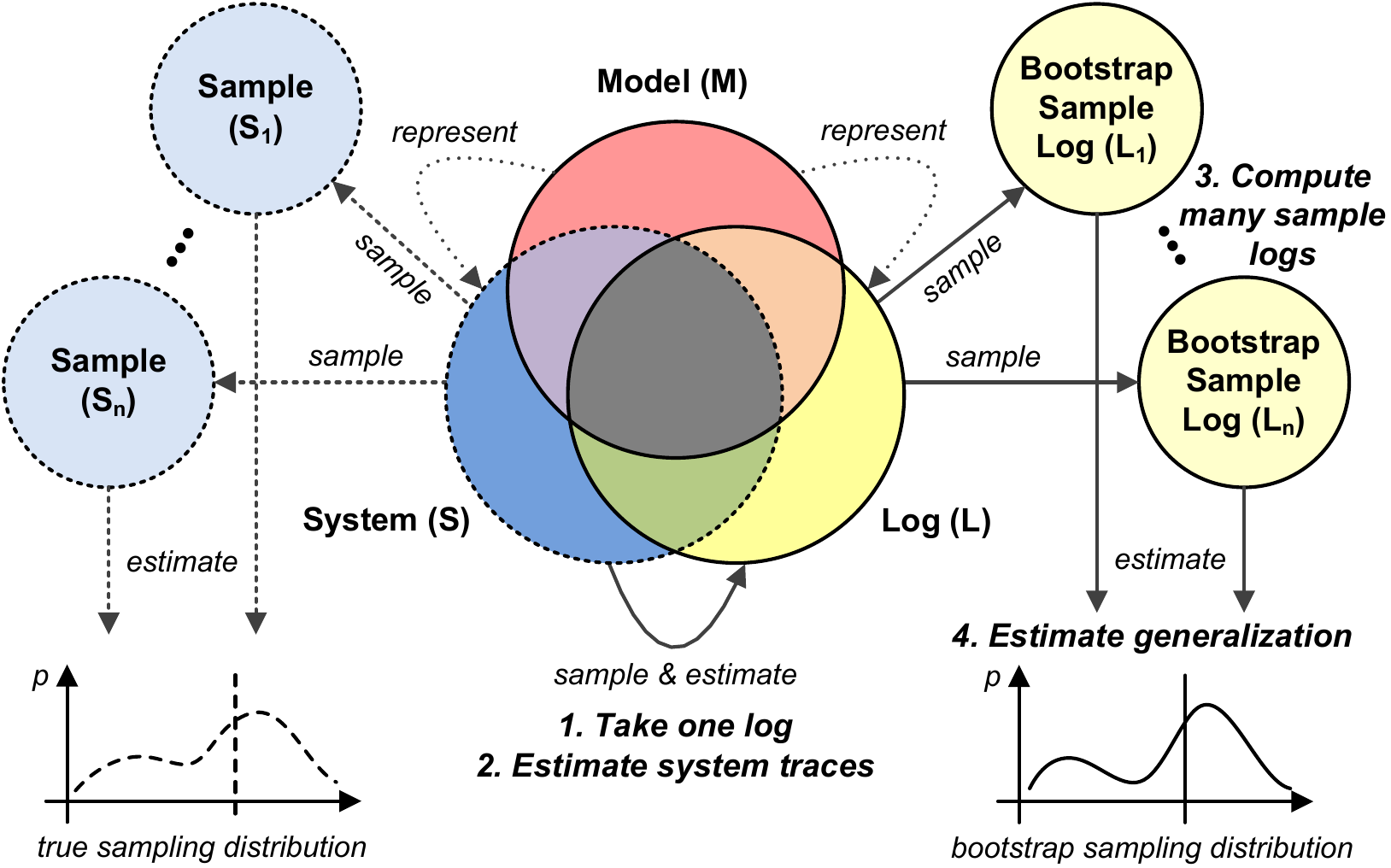}{-4}{\small Bootstrapping generalization}{-5.5}{fig:approach}

\noindent
{\Cref{fig:approach}} adapts {\cref{fig:bootstrapping}}, summarizing
the input, output, and computation of {\Cref{alg:bootstrap:gen}}.
The four generic stages
introduced above
are handled in {\cref{alg:bootstrap:gen}} as follows:
\smallskip
\begin{compactenum}[1.]
\item \emph{Take one log} of the system, as a sample of all the
traces the system can generate.
\item \emph{Estimate system traces} based on that single log.
\item \emph{Compute many samples} from the estimated system traces.
\item \emph{Estimate generalization} based on all those sample logs.
\end{compactenum}
\smallskip

\enlargethispage{\baselineskip}
\begin{algorithm}[h]
\footnotesize
\DontPrintSemicolon
\KwIn{%
Model $M\in\mathcal{M}$, log $L\in\mathcal{L}$, generalization
measure $\func{gen}{\mathcal{M} \times
\mathcal{S}}{\intervalcc{0}{1}}$, log sampling method $\mathit{lsm}
\in \mathbb{LSM}$, sample size $n\in\mathbb{N}$, and number of
samples $m\in\mathbb{N}$}
\KwOut{Estimated generalization of $M$ with respect to the system that generated $L$}
$\mathit{data}=\mset{}$\;
\For{$i \in \intintervalcc{1}{m}$}{
		sample $L_i^{*}$ of size $n$ from $L$ using $\mathit{lsm}$, \ie $L_i^*=\funcCall{lsm}{L,n}$\;
		$\mathit{data}=\mathit{data} \uplus \mset{\funcCall{gen}{M,L_i^{*}}}$\;
}
\Return{$\funcCall{average}{data}$}\;
\caption{BootstrapGeneralization($M,L,\mathit{gen},\mathit{lsm},n,m$)\label{alg:bootstrap:gen}}
\end{algorithm}

\noindent
That is, given a log $L$ (Step~1), we use $L$ itself to define the
estimated system traces (Step~2).
This decision is defensible: $L$ is a record of the system over an
extended period, and, more to the point, nothing else is known in the
scenario considered.
Step~3 appears as line~3 of {\cref{alg:bootstrap:gen}}, with
replicate logs computed using the sampling method $\mathit{lsm}$.
Next, lines~4 and~5 of {\cref{alg:bootstrap:gen}} implement Step~4 of
the generic pattern, with an estimate of the generalization
measurement computed for each sample log.
Once the individual measurements are collected, aggregation
({\emph{bagging}}) takes place at line~5.
As shown, the arithmetic mean is returned,
but other statistics can also be computed,
including confidence intervals, variance, and skewness.

\vspace{-2mm}
\subsection{Log Sampling}
\label{subsec:log:sampling}
\vspace{-1mm}

We next present two log sampling methods, that is, $\mathit{lsm}$
candidates, suitable for use in Step~2 of the generic bootstrap
scheme described in {\cref{subsec:bootstrapping}}.

There are two main forms of bootstrapping {\cite{Efron1982}}.
{\emph{Nonparametric}} bootstrapping draws samples from the data
using a ``with replacement'' methodology.
The alternative, the {\emph{parametric}} bootstrap, generates samples
using a known distribution based on parameters estimated from the
data.
Nonparametric methods reuse elements from the original sample, and
hence are only effective if the original sample is a good estimate of
the true population.
Moreover, the very essence of generalization is to measure the
model's ability to handle hitherto-unseen traces.
But nor is it clear what distribution of traces might be employed in
a parametric bootstrap for process discovery.
The first of the two log sampling techniques we explore is
nonparametric.
Let $L$ be a log, a multiset of traces, and $\funcCall{randTrace}{L}$,
a function that returns a randomly selected trace from $L$, each
chosen with probability $1/|L|$.
{\cref{alg:log:sampling:with:replacement}} describes log sampling
with replacement.

\enlargethispage{\baselineskip}
\begin{algorithm}[h]
\footnotesize
\DontPrintSemicolon
\KwIn{Log $L$, as a multiset of traces, and sample log size $n\in\mathbb{N}$}
\KwOut{Sample log $L'$}
$L'=\mset{}$\;
\lFor{$i=1$ \KwTo $n$}{$L' = L'\,\uplus\,\mset{\funcCall{randTrace}{L}}$}
\KwRet{$L'$}
\caption{LogSamplingWithReplacement($L,n$)%
\label{alg:log:sampling:with:replacement}}
\end{algorithm}

The second method we make use of is a {\emph{semiparametric}}
bootstrap, extending ideas from Theis and Darabi {\cite{Theis2020}}
(see {\cref{sec:related}}).
The semiparametric bootstrap assumes that the true population
consists of elements similar but not necessarily identical to those
in the sample; another interpretation is that a semiparametric sample
is a nonparametric sample containing a certain amount of ``noise.''
In our context, the noise is in the form of new traces; to create
them, we employ a genetic {\emph{crossover}} operator, also used in
evolutionary computation.
Two compatible {\emph{parent}} traces generate two {\emph{offspring}}
if they contain a common subtrace of some minimum length
that can become a crossover point.

Let $\subseq{t}{p}{n}$ denote the subtrace of trace $t \in
\kleenestar{\actions}$ of length $n \in \natnum$ that starts at
position $p \in \natnum$ in $t$, with $p+n-1 \leq \seqLength{t}$; and let
$\subseqid{t}{p}{n}$ {\emph{identify}} that subtrace
of $t$.
For example, $\subseqid{\trace{abbbcf}}{2}{2}$ identifies subtrace
$\trace{bb}$.
We also sometimes use underlining as a shorthand, so that
$\trace{a\underline{bb}bcf}=(\trace{abbbcf},2,2)$.
In addition, $\funcCall{prefix}{t,x}$ is the prefix of $t$ up to and
including the $x$\,th action, and $\funcCall{suffix}{t,x}$ is the
suffix of $t$ from and including that $x$\,th action.
For example, $\funcCall{prefix}{\trace{trace},3}=\trace{tra}$ and
$\funcCall{suffix}{\trace{trace},3}=\trace{ace}$.

\begin{algorithm}[h]
\footnotesize
\DontPrintSemicolon
\KwIn{Traces $t_1,t_2 \in \kleenestar{\actions}$ and length of common
subtrace $k \in \natnum$}
\KwOut{Set of all breeding sites for $t_1$ and $t_2$ for common
subtraces of length $k$}
$\mathit{sites} = \set{}$\;
\For{$p_1=1$ \KwTo $\seqLength{t_1}-k+1$}{
	\For{$p_2=1$ \KwTo $\seqLength{t_2}-k+1$}{
		\If{$\subseq{t_1}{p_1}{k}=\subseq{t_2}{p_2}{k}$}{
			$\mathit{sites} = \mathit{sites} \cup \set{(p_1,p_2)}$\;
		}
	}
}
\Return{$\mathit{sites}$}\;
\caption{BreedingSites($t_1,t_2,k$)%
\label{alg:breeding:sites}}
\end{algorithm}

Suppose that traces $t_1$ and $t_2$ share $k$ actions,
$(t_1,p_1,k)=(t_2,p_2,k)$, and that $\circ$ is a concatenation
operator.
Then, the {\emph{crossover}} operator $\otimes$ creates a new trace by
joining $t_1$ and $t_2$ across that common subtrace:
$\crossover{%
  \subseqid{t_1}{p_1}{k}
  }{%
  \subseqid{t_2}{p_2}{k}%
  }{%
  \concat{%
    \funcCall{prefix}{t_1,p_1+k-1}%
    }{%
    \funcCall{suffix}{t_2,p_2+k}%
    }
  }%
$.
For example, traces {\trace{abbcf}} and {\trace{abbbbcf}} are
obtained from {\trace{abbbcf}} via self-crossover, with {\texttt{bb}}
appearing at the {\emph{breeding sites}} $p_1=2$ and $p_2=3$, yielding
$\crossover{\trace{a\underline{bb}bcf}}{\trace{ab\underline{bb}cf}}{\trace{abbcf}}$,
and
$\crossover{\trace{ab\underline{bb}cf}}{\trace{a\underline{bb}bcf}}{\trace{abbbbcf}}$.
Two traces might have multiple breeding sites, with the count
determined by the traces and the value of $k$.
{\Cref{alg:breeding:sites}} identifies all breeding sites for two
input traces.
For example, {\trace{adeef}} and {\trace{adefabcfadef}} have six
$k=2$ breeding sites:
$\set{\pair{1}{1}, \pair{1}{9}, \pair{2}{2},
\pair{2}{10}, \pair{4}{3}, \pair{4}{11}}$.

\begin{algorithm}[h]
\footnotesize
\DontPrintSemicolon
\KwIn{Logs $L_1$ and $L_2$, as multisets of traces, length of common subtrace $k \in \natnum$, and breeding probability $p \in \intervalcc{0}{1}$}
\KwOut{Log $L'$ of traces that result from breeding $L_1$ and $L_2$}
$L'=\mset{}$\;
\For{$i=1$ \KwTo $\ceil{{\cardinality{L_1}}/{2}}$}{
	$t_1 = \funcCall{randTrace}{L_1}$\;
	$t_2 = \funcCall{randTrace}{L_2}$\;
	$\mathit{sites} = \funcCall{\text{BreedingSites}}{t_1,t_2,k}$\;
	\eIf{$\funcCall{rand}{} < p$ {\bf and} $\mathit{sites}\neq\mset{}$}{
		select a random pair $(p_1,p_2)$ from $\mathit{sites}$\;
		$L' = L'\,\uplus\mset{\crossoverop{\subseqid{t_1}{p_1}{k}}{\subseqid{t_2}{p_2}{k}},\crossoverop{\subseqid{t_2}{p_2}{k}}{\subseqid{t_1}{p_1}{k}}}$\;
	}{
		$L' = L'\,\uplus\mset{t_1,t_2}$\;
	}
}
\KwRet{$L'$}
\caption{LogBreeding($L_1,L_2,k,p$)%
\label{alg:log:sampling:with:breeding}}
\end{algorithm}

\enlargethispage{\baselineskip}
In terms of a system or model, each possible candidate crossover site
represents a ``hyper jump'' between pairs of states that share a
common $k$-action context.
\updated{We do not claim that} all systems actually behave in this
way; but {\cref{lem:crossover:reliable:DFAs}}, below, shows that some
interesting classes of systems do.
A noteworthy property of the crossover operator is
that it allows loops to be inferred if traces
that include the loop appear in the log.
For example, in {\cref{fig:example:system}}, the state labeled
{\action{b}} is the location of a loop of length one, with
both of {\trace{ab}} and {\trace{bb}} as $k=2$ contexts; and, as
already noted, the crossover operator can spawn both {\trace{abbcf}}
and {\trace{abbbbcf}} if {\trace{abbbcf}} is available in the log.

\begin{algorithm}[h]
\footnotesize
\DontPrintSemicolon
\KwIn{Log $L$, a multiset of traces, and
sample log size $n \in \natnum$.
The number of log generations $g \in
\natnum$, the common subtrace length $k \in \natnum$, and the breeding
probability $p \in \intervalcc{0}{1}$ are assumed to be constants}
\KwOut{Sample log $L'$}
$G[0]=L$\;
\lFor{$i=1$ \KwTo $g$}{$G[i]=\funcCall{\text{LogBreeding}}{L,G[i-1],k,p}$}
$L'= \funcCall{\text{LogSamplingWithReplacement}}{\cup_{i=0}^{g}{G[i]}, n}$\;
{\bf return }$L'$\;
\caption{LogSamplingWithBreeding($L,n$)%
\label{alg:log:sampling:with:breeding:n}}
\end{algorithm}

{\Cref{alg:log:sampling:with:breeding,alg:log:sampling:with:breeding:n}}
crystallize these ideas, assuming that $\funcCall{rand}{}$ returns a
uniformly distributed value in $\intervalcc{0}{1}$.
In {\cref{alg:log:sampling:with:breeding}}, traces are chosen from
each of $L_1$ and $L_2$, and then, with some probability $p$, checked
for $k$-overlaps, and permitted to breed.
If they do breed, their offspring are added to the output set; if
they do not, the strings themselves are added.
That process iterates until $L'$ contains $\approx|L_1|$ traces.
{\Cref{alg:log:sampling:with:breeding:n}} then adds the notion of
{\emph{generations}}, with the output log $L'$ of size $n$ a random
selection across traces formed during $g$ generations of breeding,
where the $i$\,th generation arises when the original log $L$ is bred
with the $i-1$\,th generation.
{\Cref{alg:log:sampling:with:breeding:n}} thus provides a
semiparametric $\mathit{lms}$ sampler that can, like
{\cref{alg:log:sampling:with:replacement}}, be used for
bootstrapping.

\vspace{-2mm}
\subsection{Generalization Measures}
\label{subsec:measures}
\vspace{-1mm}

We now present two measures that quantify the ability of a model to
represent a system.

As noted in~\cref{subsec:generalization}, Buijs et
al.~\cite{Buijs2014} suggest that model-system recall be used to
measure generalization.
However, that proposal has two limitations.
First, the measure is of only limited utility when models can
describe infinite collections of traces, as cardinality measures over
sets become problematic.
Second, given a model $M$ and system $S$, but where $M \not\subseteq
S$, the suggested calculation is indeterminate.
The first limitation can be resolved by replacing the cardinality
measure over sets with $\funcCall{ent}{\cdot}$, a measure inspired by
the topological entropy of a potentially infinite
language~\cite{Polyvyanyy2020b}.
The result is a measure referred to as the {\emph{coverage of $M$
with $S$}}, and, in essence, is the model-system recall instantiated
with the entropy as an estimation of cardinality:
\eqntopspace
\begin{equation}
\label{eq:model:system:recall:entropy}
\funcCall{ModelSystemRecall}{M,S} = \frac{\funcCall{ent}{M \cap S}}{\funcCall{ent}{M}}\,.
\eqnbotspace
\end{equation}
By analogy, we now suggest addressing the second limitation by
considering model-system precision as a second aspect that
characterizes the generalization of the model\footnote{Both can be
computed using {\textsl{Entropia}} {\cite{Polyvyanyy2020a}}.
Recall is specified by the {\texttt{-emr}} option, and precision by
{\texttt{-emp}}.
Languages are compared based on exact matching of constituent traces,
based on models and systems provided as Petri nets.}:
\eqntopspace
\begin{equation}
\label{eq:model:system:precision:entropy}
\funcCall{ModelSystemPrecision}{M,S} = \frac{\funcCall{ent}{M \cap S}}{\funcCall{ent}{S}}.
\eqnbotspace
\end{equation}

Model-system precision and recall can both be reported, or a single
blended value -- their harmonic mean, for example -- can be computed.
We postpone discussion of which approach is preferable to future
work.
The entropy-based model-log measures of precision and recall satisfy
all the desired properties for the corresponding class of
measures~\cite{Syring2019}, making it interesting to study how these
measures perform, in terms of generalization
properties~\cite{Aalst2018}, when comparing the traces of the model
and system.

\vspace{-2mm}
\subsection{Consistency}
\label{subsec:consistency}
\vspace{-1mm}
\enlargethispage{\baselineskip}

Next, we show that our estimator of generalization is consistent for
systems captured as DFGs, which are graphs of actions commonly
used by industry to describe process models~\cite{Aalst2016}, making
it reasonable to assume that the unknown systems they correspond to
are also captured as DFGs.
{\Cref{fig:1}} shows two DFGs.

\begin{define}{DFG}{def:FDG} A
{\emph{directly-follows graph}} (DFG) is a tuple $(\Phi, \Psi, \phi,
\psi, i, o)$; with $\Phi \subseteq \actions$ a set of {\emph{actions}};
$\Psi \subseteq ((\Phi \times \Phi) \cup (\set{i} \times \Phi) \cup
(\Phi \times \set{o}))$ a {\emph{directly-follows relation}};
$\func{\phi}{\Phi \cup \set{i,o}}{\natnumwithzero}$ an
{\emph{action frequency function}};
$\func{\psi}{\Psi}{\natnumwithzero}$ an {\emph{arc frequency
function}}; and $i \not\in \actions$ and $o \not\in \actions$ the
{\emph{input}} and the {\emph{output}} of the graph.
\end{define}

\noindent
\updated{We define the semantics of a DFG via a mapping to a finite automaton~{\cite{PolyvyanyyMG20}}.}

\begin{define}{DFA}{def:FDA} 
\updated{A {\emph{deterministic finite automaton}} (DFA) is a tuple
$\tuple{Q,\actions,\delta,q_0,A}$, with $Q$ a finite set of
{\emph{states}}; $\actions$ a finite set of \emph{actions}; $\func{\delta}{Q
\times \actions}{}$ $Q$ the {\emph{transition function}}; $q_0 \in Q$
the {\emph{start state}}; and $A \subseteq Q$ is the set of {\emph{accepting states}}.}
\end{define}

\noindent
A sequence of actions is a trace of a DFA if the DFA
accepts that sequence of actions.
A DFA is {\emph{stable}} if $\forall (q_1,\lambda,q_2) \in \delta \wedge
\forall (q'_1,\lambda',q'_2) \in \delta : ( ( \lambda = \lambda' )
\implies ( q_2 = q'_2 ) )$.
A DFG $(\Phi,\Psi,\phi,\psi,i,o)$ gives rise to a DFA
$(\Phi \cup \set{i,o},\Phi \cup \set{o},\delta,i,o)$,
with
$\delta = \{ (s,t,t) \in (\set{i} \cup \Phi)\times(\Phi \cup
\set{o})\times(\Phi \cup \set{o}) \mid (s,t) \in \Psi\}$
that is guaranteed to be stable. 

\begin{lem}{Stable DFAs}{lem:stable:DFAs}
A DFA of a DFG is stable.
\end{lem}

\noindent
\updated{Indeed, an occurrence of an action is always followed by the same opportunities for future actions; and hence, any offspring that result from the
crossover of two traces of a stable DFA are also traces of the DFA.}

\begin{lem}{Trace crossover}{lem:crossover:reliable:DFAs} If $t_1,t_2
\in \kleenestar{\actions}$ are traces of a stable DFA
and if $t = (t_1,p_1,1)
\otimes (t_2,p_2,1)$, for $p_1,p_2 \in \natnum$, then $t$ is accepted
by the DFA.
\end{lem}

\noindent
\emph{Proof sketch.}
By definition,
$t=%
\concat{%
\funcCall{prefix}{t_1,p_1}%
}{%
\funcCall{suffix}{t_2,p_2+1}%
}$, and hence
the elements in $t_1$ and $t_2$ at positions $p_1$ and
$p_2$ are instances of the same action.
As the DFA is stable, $\funcCall{prefix}{t_1,p_1}$ and
$\funcCall{prefix}{t_2,p_2}$ lead to the same state $q$ in the DFA; and
because $\funcCall{suffix}{t_2,p_2+1}$ leads from $q$ to an accept state,
$t$ must also be accepted by the DFA.
\hfill\ensuremath{\blacksquare}
\medskip

\noindent
If two traces share a crossover of any length, there must also be a
crossover of length one that results in the same offspring pair.
Consequently, a log sample that results from
{\cref{alg:log:sampling:with:breeding:n}} for an input log composed
of traces from a system that is a DFG will also contain valid traces.
Such a log sample estimates the system at least as well
as the original log.
One further condition is then sufficient to allow our main result.

\enlargethispage{\baselineskip}
\begin{thm}{Bootstrapping DFAs}{thm:bootstrapping:DFAs}{\quad\\} Let
$L$ be a set of traces from a stable DFA describing a language $L^*$,
$L \subseteq L^*$, such that each subtrace of length two of any trace
in $L^*$ is also a subtrace of some trace in $L$.
Then $L'$ is a log of the DFA with $L \subseteq L'$ and $L' \subseteq
L^*$ iff $L'$ can result from log sampling with breeding
(\cref{alg:log:sampling:with:breeding:n}) for input log $L$ and
common subtrace length $k=1$.
\end{thm}

\noindent
\emph{Proof sketch.}
($\Rightarrow$) 
If $t \in L'$ is not a crossover of two sequences in $L'$
then $t$ is a trace of the DFA (base case).
Otherwise, let $t=(t_1,p_1,1) \otimes (t_2,p_2,1)$, where $t_1$ and
$t_2$ are traces of the DFA.
As the DFA is stable, $t$ is a trace of the DFA, and the action at
position $p_1+1$ in $t$ is taken from the state of the DFA reached
after the action at position $p_1$.
\\
\noindent
($\Leftarrow$)
Let $t \in L'$, and consider two cases.
(i) If $t \in L$, then $t$ is a trace of the DFA.
(ii) Suppose $t \not\in L$.
But $\funcCall{prefix}{t,0}$ is a computation of the DFA, and if
$\funcCall{prefix}{t,k}$, $k < \seqLength{t}$ is a computation of the
DFA, then $\funcCall{prefix}{t,k+1}$ is also computation of the DFA,
via two subcases.
(ii.a) If $t=(t_1,k,1) \otimes (t_2,m,1)$, $m \in
\natnum$, $t_1,t_2 \in L'$ it follows
(the DFA is stable) that $\funcCall{prefix}{t,k+1}$ is a
computation of the DFA.
Indeed, $t_1$ and $t_2$ are traces of the DFA, shown by structural
induction on the hierarchy of crossovers over the sequences in $L'$,
and the last action in the prefix is taken from the same state of the
DFA.
(ii.b) Otherwise, $\funcCall{prefix}{t,k+1}$ is a prefix of some
trace of the DFA and, thus, is its computation, implying that $t$
leads to an accept state, as its last action is the last action of
some trace in $L$.
\hfill\ensuremath{\blacksquare}
\smallskip

\noindent
Hence, the larger the bootstrapped samples of a DFG log that are
generated, the better the estimate of the system -- meaning that
bootstrap generalization (\cref{alg:bootstrap:gen}) instantiated with
the entropy-based model-system measures
(\cref{eq:model:system:recall:entropy,eq:model:system:precision:entropy})
is consistent, a consequence of the monotonicity property of the two
model-system measures~\cite{Polyvyanyy2020b}.

\vspace{-2mm}
\subsection{Example}
\label{subsec:example}
\vspace{-1mm}

Consider again the running example of
{\cref{subsec:process:discovery}}.
For the languages $M$ and $S$ described by the DFGs of
{\cref{fig:example:process:model}} and {\cref{fig:example:system}},
$\funcCall{ModelSystemRecall}{M,S}=\num[round-precision=3]{0.867}$ and
$\funcCall{ModelSystemPrecision}{M,S}=\num[round-precision=3]{0.867}$, noting that
precision and recall are the same if the complexity of the system and
model languages is the same {\cite{Polyvyanyy2020b}}.

\enlargethispage{\baselineskip}
Assuming now that $S$ is unknown, we apply {\cref{alg:bootstrap:gen}}
(BootstrapGeneralization) to estimate the corresponding measurements,
with parameters: input model $M$; the log $L$ of 66 traces presented
in {\cref{subsec:process:discovery}}; the generalization measures of
{\cref{eq:model:system:recall:entropy,eq:model:system:precision:entropy}}
($\mathit{gen}$); log sampling with breeding as described by
{\cref{alg:log:sampling:with:breeding:n}} ($\mathit{lsm}$); sample
log sizes of $n=\num{100000}$ and $\num{1000000}$ traces;
$m=\num{100}$ log replicates; $g=\num{10000}$ log generations;
breeding sites of length $k=2$; and a breeding probability of
$p=\num[round-precision=1]{1.0}$.
The estimation process yielded model-system precision and recall
measurements of $\num[round-precision=3]{0.892209}$ and
$\num[round-precision=3]{0.912492}$ (for $n=\num{100000}$), and of
$\num[round-precision=3]{0.897187}$ and
$\num[round-precision=3]{0.908485}$ ($n=\num{1000000}$).
In contrast, the original log $L$ of 66 traces does not provide a
good representation of the system, with model-log precision and
recall of $\num[round-precision=3]{0.791}$ and
$\num[round-precision=3]{0.935}$, respectively.
The two computations took $\num{457}$ and $\num{575}$ seconds,
respectively,
on a commodity laptop running Windows 10, Intel(R)
Core(TM) i7-7500U CPU @ 2.70GhZ and 16GB of RAM.

\begin{table}[h]
\small
\caption{\small Precision and recall estimates via bootstrapping, plus the
number of distinct traces per replica, together with 95\% confidence
intervals, using $m=\num{100}$ replicates throughout: (a) varying
$n$, the number of traces per replicate, with $g=\num{10000}$
generations held constant; and (b) varying $g$, with $n=\num{10000}$
held constant.
The confidence intervals for precision and recall are 
for the estimated values considering the input parameters and, thus, 
might \emph{not} include the true values.
}
\label{tbl:sample:results}
\centering
\begin{tabular}{c@{\hspace{1em}}c}
(a) & (b)
\\
\begin{tabular}{
	c
	*{2}{
		S[round-precision=2,table-format=1.2]
		@{$\,\pm\,$}
		S[round-precision=2,table-format=1.2]
	}
	S[round-precision=0,table-format=3.0]
	@{$\,\pm\,$}
	S[round-precision=1,table-format=1.1]
}
\toprule
$n$
	& \multicolumn{2}{c}{\emph{precision}}
		& \multicolumn{2}{c}{\emph{recall}}
			& \multicolumn{2}{c}{\emph{traces}}
\\
\midrule
\num{100}
	& 0.8345659561 & 0.0013393
		& 0.9518126348 & 0.0023022
			& 11.91 & 0.3249874
\\
\num{1000}
	& 0.8626786708 & 0.0012330
		& 0.9298854831 & 0.0018554
			& 27.69 & 0.6082271
\\
\num{10000}
	& 0.8814277959 & 0.0007629
		& 0.9187321500 & 0.0007010
			& 56.54 & 0.7425113
\\
\num{100000}
	& 0.8922092580 & 0.0003874
		& 0.9124916496 & 0.0004989
			& 107.30 & 1.0309237
\\
\num{1000000}
	& 0.8971867148 & 0.0002738
		& 0.9084846702 & 0.0003349
			& 166.22 & 1.3751049
\\
\bottomrule
\end{tabular}
&
\begin{tabular}{
	c
	*{2}{
		S[round-precision=2,table-format=1.2]
		@{$\,\pm\,$}
		S[round-precision=2,table-format=1.2]
	}
	S[round-precision=0,table-format=3.0]
	@{$\,\pm\,$}
	S[round-precision=1,table-format=1.1]
}
\toprule
$g$
	& \multicolumn{2}{c}{\emph{precision}}
		& \multicolumn{2}{c}{\emph{recall}}
			& \multicolumn{2}{c}{\emph{traces}}
\\
\midrule
\num{100}
	& 0.8720707631 & 0.0010986
		& 0.9206088554 & 0.0009082
			& 41.62 & 0.8106451
\\
\num{1000}
	& 0.8802371562 & 0.0008274
		& 0.9192678251 & 0.0006669
			& 53.66 & 0.7686004
\\
\num{10000}
	& 0.8814277959 & 0.0007629
		& 0.9187321500 & 0.0007010
			& 56.54 & 0.7425113
\\
\num{100000}
	& 0.8820194857 & 0.0007065
		& 0.9191841044 & 0.0005925
			& 56.82 & 0.7346096
\\
\num{1000000}
	& 0.8807455009 & 0.0006947
		& 0.9184879627 & 0.0006232
			& 55.90 & 0.6916909
\\
\bottomrule
\end{tabular}
\\
\end{tabular}
\end{table}

\noindent
{\Cref{tbl:sample:results}} shows other values generated by
bootstrapping.
The simplicity of the example configuration -- with just a
handful of distinct traces in $L$, and hence a very limited range of
$k=2$ breeding sites -- means that the number of distinct traces per
replicate log grows relatively slowly.
However, as the traces of the log contain all the subtraces of length
two that can be found in traces in the language of the system
{\updated{it is guaranteed (\cref{thm:bootstrapping:DFAs})}} that the
larger the bootstrapped logs become, the more complete the coverage of
the system and, consequently, {\updated{the more
accurate the estimated generalization}}.


\vspace{-2mm}
\section{Evaluation}
\label{sec:evaluation}
\vspace{-1mm}
\enlargethispage{\baselineskip}

\subsection{Data and Experimentation}
\label{subsec:data:and:experimentation}

{\Cref{alg:bootstrap:gen,alg:log:sampling:with:replacement,alg:breeding:sites,alg:log:sampling:with:breeding,alg:log:sampling:with:breeding:n}}
were implemented\footnote{See
\url{https://github.com/lgbanuelos/bsgen} for public software.}
and used to demonstrate the feasibility of our approach
when used in (close to) industrial settings.
A set of {\num{60}} DFGs shared with us by Celonis SE
(\url{https://www.celonis.com}) was then used as a library of ground truth
systems {\updated{\cite{PolyvyanyyMG20,Alkhammash2020}}}.
Those reference DFGs were generated from three source logs (Road
Traffic Fine Management Process, RTFMP
{\cite{DeLeoniM.Massimiliano2015}}, Sepsis Cases
{\cite{Mannhardt2016a}}, and BPI Challenge 2012
{\cite{VanDongen2012}}); two different discovery techniques (denoted
``PE'' and ``VE''); and ten combinations of parameter settings
(denoted ``01'' to ``10'').

For each of the {\num{60}} DFGs, we constructed a log of
{\num{100}} traces by taking ``random walks'' through its states.
Commencing at the start vertex, the first {\emph{context}}, one
action was chosen uniformly randomly from the edges available, and
the context switched to the destination of that edge.
That process was iterated until the final state of the system was
reached as the context (every non-final state in these models has at
least one outward edge), thereby generating one trace in the
corresponding log.

Next, from each of the {\num{60}} generated logs, we discovered a
process model using the Inductive Mining algorithm with a noise
threshold of $0.8$ {\cite{Leemans2013}}.
In this controlled experimental setting, in which all of system
($S$), log ($L$), and discovered model ($M$) are known, we have the
ability to compute true model-system precision and recall
({\cref{eq:model:system:precision:entropy,eq:model:system:recall:entropy}}),
that is, the ground truth generalization of the derived model.

Then we ``forget'' about the ground truth system, and estimate the
same measurements using {\cref{alg:bootstrap:gen}}, invoked on each
combination of derived model $M$ and log $L$, in conjunction with:
model-system precision and recall measures ($\mathit{gen}$); log
sampling with trace breeding (\cref{alg:log:sampling:with:breeding}
as $\mathit{lsm}$); a sample log size of $n=\num{100000}$;
$m=\num{50}$ log replicates; $g=\num{10000}$ log generations; a
common subtrace length of $k=\num{2}$; and a breeding probability of
$p=1.0$.
All computation was on a Linux server with Intel(R) Xeon(R) Processor
(Cascadelake), {\num{32}} cores @ 2.0GHz each, and {\num{288}}GB of
memory.

\vspace{-2mm}
\subsection{Results}
\label{subsec:results}
\vspace{-1mm}

A subset of results is shown in {\cref{tbl:results}}, covering twelve
systems (three original processes, the ``PE'' and ``VE'' discovery
mechanisms, and the ``04'' and ``07'' parameter settings), with each
row showing data for a single ground truth system.
{\updated{The columns ``{\emph{model-system}}'' and
``{\emph{model-log}}'' report true model-system precision and recall
and the corresponding model-log values; and the columns
``{\emph{bootstrapped generalization}}''}} give estimated
model-system precision and recall computed via the new bootstrapping
process, together with $95$\% confidence intervals.
\updated{%
All of the bootstrapped values are closer to the true generalization
values than the corresponding model-log values, confirming the
applicability of the new approach.
For example, in the first row in Table~\ref{tbl:results} the true
value of model-system precision, which as discussed in
{\cref{subsec:measures}} is used as a measure of generalization,
is $0.60$.
The precision between that model and the log is $0.48$, while the
bootstrapped precision is equal to $0.55 \pm 0.00$, better
approximating $0.60$.}

\begin{table*}[t]
\vspace{-2mm}
\caption{\small True model-system precision and recall, model-log
precision and recall, and estimated precision and recall via
bootstrapping, plus the number of distinct traces per replica,
together with 95\% confidence intervals, see the text for
configuration details.
}
\label{tbl:results}
\centering
\sisetup{
group-separator={,},
round-mode=places,
round-precision=2,
table-format=1.2,
}
\begin{tabular}{@{}
	cl
	*{2}{S[round-precision=0,table-format=2.0]}
	|
	*{2}{S}
	|
	*{2}{S}
	|
	*{2}{S@{$\,\pm\,$}S[table-format=1.2,round-precision=2]}
	S[table-format=4.0,round-precision=0]
	@{$\,\pm\,$}
	S[table-format=2.0,round-precision=0]
	@{}
}
\toprule
\multicolumn{4}{c|}{\textit{system}}
	& \multicolumn{2}{c|}{\textit{model-system}}
	& \multicolumn{2}{c|}{\textit{model-log}}
	& \multicolumn{6}{c}{\textit{bootstrapped generalization}}
\\
\midrule
	& \textit{name}      
	& \textit{nodes}
	& \textit{edges}
	& \textit{prec.} 
		& \textit{recall}
	& \textit{prec.}
		& \textit{recall}
	& \multicolumn{2}{c}{\textit{precision}}
	& \multicolumn{2}{c}{\textit{recall}}
	& \multicolumn{2}{c}{\textit{traces}}
\\
\midrule
	1 &
PE BPI Chall. 04
	& 16 & 26			
	& 0.5952258 & 1.0000000		
	& 0.4821084 & 0.9992454		
	& 0.5526756 & 0.0003233		
	& 0.9997662 & 0.0002895		
	& 3540.5217391 & 13.4801883	
\\
	2 &
PE BPI Chall. 07
	& 25 & 48			
	& 0.2509794 & 1.0000000		
	& 0.1748664 & 0.9998886		
	& 0.1942679 & 0.0001035		
	& 1.0000256 & 0.0000549		
	& 2488.6938776 & 9.9697141	
\\
	3 &
VE BPI Chall. 04
	& 16 & 34			
	& 0.5656396 & 0.9999817		
	& 0.3840598 & 1.0000164		
	& 0.4649622 & 0.0001136		
	& 0.9999981 & 0.0000153		
	& 2383.8800000 & 10.4894070	
\\
	4 &
VE BPI Chall. 07
	& 20 & 57			
	& 0.4484944 & 1.0000000		
	& 0.2306200 & 0.9998963		
	& 0.2819579 & 0.0001623		
	& 0.9999972 & 0.0000202		
	& 3088.5600000 & 9.5432105	
\\
	5 &
PE RTFMP 04
	& 12 & 24			
	& 0.4429498 & 1.0000000		
	& 0.3807623 & 0.9999801		
	& 0.4294444 & 0.0000551		
	& 0.9999957 & 0.0000084		
	& 921.7600000 & 4.3389483	
\\
	6 &
PE RTFMP 07
	& 13 & 54			
	& 0.4566679 & 1.0000000		
	& 0.2610725 & 1.0000000		
	& 0.3330580 & 0.0002440		
	& 1.0000000 & 0.0000000		
	& 2521.2000000 & 9.2343984	
\\
	7 &
VE RTFMP 04
	& 10 & 29			
	& 0.5970821 & 1.0000000		
	& 0.3988987 & 0.9999999		
	& 0.4881633 & 0.0003195		
	& 1.0000007 & 0.0000006		
	& 2163.9800000 & 9.5896484	
\\
	8 &
VE RTFMP 07
	& 13 & 58			
	& 0.4750158 & 1.0000000		
	& 0.2477764 & 0.9999989		
	& 0.3331886 & 0.0001883		
	& 1.0000000 & 0.0000001		
	& 3860.2400000 & 13.3253502	
\\
	9 &
PE Sepsis Cas. 04
	& 15 & 35			
	& 0.4712458 & 1.0000000		
	& 0.2280462 & 0.9998199		
	& 0.2851726 & 0.0001059		
	& 0.9996701 & 0.0002815		
	& 5195.5000000 & 12.8027475	
\\
	10 &
PE Sepsis Cas. 07
	& 17 & 64			
	& 0.7054537 & 1.0000000		
	& 0.2435910 & 1.0001455		
	& 0.3170748 & 0.0001888		
	& 1.0000000 & 0.0000000		
	& 5405.5800000 & 18.1526596	
\\
	11 &
VE Sepsis Cas. 04
	& 12 & 23			
	& 0.7041809 & 1.0000000		
	& 0.5350890 & 1.0000150		
	& 0.5829114 & 0.0002158		
	& 1.0001295 & 0.0001390		
	& 201.6400000 & 1.6198713	
\\
	12 &
VE Sepsis Cas. 07
	& 13 & 53			
	& 0.6582486 & 1.0000000		
	& 0.2815729 & 1.0000031		
	& 0.3622550 & 0.0002244		
	& 0.9999989 & 0.0000031		
	& 4777.0000000 & 14.4505559	
\\
\bottomrule
\end{tabular}
{\footnotesize The complete version of the table is available at
{\url{http://go.unimelb.edu.au/52gi}}.}
\vspace{-2mm}
\end{table*}

\updated{The small systems perform consistently better.
This suggests the need for further trace breeding mechanisms that
target large systems.
For example, the bootstrapped precision for the Sepsis Cases log
(discovery technique ``VE'' and parameter ``07'', in row 12) is $0.36
\pm 0.00$, which is closer to the true model-system
precision than is $0.28$, the model-log precision,
but still notably different from~$0.66$.
Note that in this case the DFG has $53$ edges, twice as
many as the example in the first row of the table.}
{\updated{Avenues for further work thus include assessing different
log sampling mechanisms in terms of their}} accuracy (sampling traces
supported by the system); their velocity (sampling accurate traces
quickly); and their stability (sampling traces that lead to
consistent measurements).
For example, the method used in {\cref{tbl:results}} is stable, as
evidenced by the small confidence intervals of the estimates, but is
slow, in that it requires many generations to breed relatively small
numbers of new traces.

\vspace{-2mm}
\subsection{Threats to Validity}
\label{subsec:validity}
\vspace{-1mm}

\updated{%
Several threats to validity are worth mentioning.
Firstly, the discovered models were accepted as ground truth systems.
These models were discovered by process mining experts independently,
without the involvement of the authors of this paper.
Nevertheless, they may not represent actual systems accurately.
An obvious next step is thus to extend the experiments to datasets
that include both the true system models and also logs induced from
them.
At the moment, such datasets are not available to the process mining
research community.
Secondly, the collection of $60$ system models used in the evaluation
is not representative of the full spectrum of possible systems;
indeed, all of the recall measurements ended up being $1.0$.
They also come from a limited set of domains, namely, healthcare,
loan application, and road traffic management.
Hence, while the results confirm the consistency of the estimation
approach shown in {\cref{subsec:consistency}}, they also demonstrate
different behaviors -- notably, convergence rates -- for different
(classes of) systems.
Further experiments with real-world and synthetic systems and logs
will help to understand such properties better.}


\vspace{-3mm}
\section{Related Work}
\label{sec:related}
\vspace{-2mm}
\enlargethispage{\baselineskip}

Generalization is perhaps the least-studied quality criterion of
discovered process models, with just a small number of measures
proposed; we now briefly survey those.

Given a function that maps log events onto states in which they
occur, {\emph{alignment generalization}} {\cite{Aalst2012}} counts
the number of visits to each state, and the number of different
events that occur in each state.
If states are visited often and the number of events observed from
them is low, it is unlikely that further events will arise, and
generalization is good.
van der Aalst et al.~\cite{Aalst2012} also propose different forms of
cross-validation methodology, including a ``leave one out'' approach.
The bootstrap method provides more general sample reuse
{\cite{Efron1982}}, and, as discussed in
{\cref{subsec:bootstrapping}}, estimates the population using the
sample, and computes fresh samples from the estimated population,
rather than using the single sample to both train the prediction
model and to assess the prediction error.

{\emph{Weighted behavioral generalization}} {\cite{Broucke2014}}
measures the ratio of allowed generalizations to the allowed plus
disallowed generalizations.
Allowed and disallowed generalizations are determined based on
``weighted negative events,'' which capture the fact that the event
cannot occur at some position in a trace.
The event weight reflects the likelihood of the event being observed
in future traces of the system.
The more disallowed generalizations that are identified, the lower
the generalization.

{\emph{Anti-alignment based generalization}}~\cite{Dongen2016}
promotes models that describe traces not in the log, without
introducing new states.
The underlying intuition is that the log describes a significant
share of the state space of the system, and that future system traces
may trigger fresh actions from known states, but not fresh states.
It is implemented using a leave one out cross-validation strategy and
``anti-alignments,'' traces from the model that are as different as
possible to those in the log.

The {\emph{adversarial system variant
approximation}}~\cite{Theis2020} uses the log's traces to train a
sequence generative adversarial network (SGAN) that approximates the
distribution of system traces, and then employs a sample of traces
induced by the SGAN to represent the system's behavior.
Generalization is then measured using standard approaches.
Trained SGANs can also be incorporated into our bootstrap-based
method to obtain a parametric bootstrap strategy, an option we will
explore in future work.

Other approaches to measuring generalization have also been
proposed~\cite{Aalst2018,Buijs2014}.
However, they are only partially able to analyze models with loops.
van der Aalst~\cite{Aalst2018} lists ten properties (including three
that are subject to debate) that a generalization measure should
satisfy; and it has been shown~\cite{Aalst2018,Syring2019} that
existing measures don't satisfy the seven properties that are
agreed~\cite{Aalst2012,Broucke2014,Dongen2016,Aalst2018}.
Moreover, Janssenswillen et al.~\cite{Janssenswillen2017} show that
existing generalization measures assess different
phenomena~\cite{Aalst2012,Buijs2014,Broucke2014}.
As the instantiations of the generalization estimator discussed and
evaluated in this work rely on the entropy-based precision and recall
measures~\cite{Polyvyanyy2020b}, which were shown to satisfy all the
desired properties for the corresponding classes of
measures~\cite{Syring2019}, it is interesting to study the properties
of our generalization estimators.
However, properties of the estimators must be studied in the limit
(as the input grows and the estimators converge), which requires
adjustments of the original properties.
We will do that as future work.

An experiment with synthetic models and simulated logs analyzed
whether existing model-log precision and recall, and generalization
measures can be used as estimators of model-system precision and
recall~\cite{Janssenswillen2018}.
The experiment measured model-system and model-log properties, and
performed statistical analysis to establish relationships between
them.
The reported results indicate that using currently available methods,
it is ``nearly impossible to objectively measure the ability of a
model to represent the system.''
In our work, instead of relating model-system and model-log
measurements, we use the bootstrap method to estimate the entire
behavior of the system from its log, and then measure model-system
properties using the model and the estimated behavior of the system.
Under reasonable assumption, our estimator of generalization is
consistent.

Also related to the problem of measuring generalization of a
discovered model is establishing the {\emph{rediscoverability}} of a
process discovery algorithm, that is, identifying conditions under
which it constructs a model that is behaviorally equivalent to the
system and describes the same set of traces~\cite{Aalst2016}.
Such conditions usually address both the class of systems and the
class of logs for which rediscoverability can be assured.
For example, the Inductive Mining algorithm guarantees
rediscoverability for the class of systems that are captured as
block-structured process models~\cite{Polyvyanyy2012b} without
duplicate actions and in which it is not possible to start a loop
with an activity the same loop can also end with~\cite{Leemans2018}.
In contrast to rediscoverability guarantees, we study the problem of
measuring how well a discovered model describes an unknown original
system.

\vspace{-2mm}
\section{Conclusion}
\label{sec:conclusion}
\vspace{-1mm}

We presented a bootstrap-based approach for estimating the
generalization of models discovered from logs, parameterized
by a generalization measure defined over known systems, and a log
sampling method.
An instantiation using entropy-based generalization and log sampling
based on $k$-overlap breeding of traces is shown to be consistent for
the class of systems captured as DFGs.
Thus, the larger the constructed samples and the more samples
get bootstrapped, the more accurate the estimation of the
generalization is.
Our evaluation confirmed the approach's feasibility in industrial
settings.

This work marks a first step in a study of the applicability of
bootstrap methods for estimating generalization, and can be extended
in several ways.
In future work we will seek to develop an unbiased estimator, that
is, an estimator with no difference between the expected value of the
estimation and the true value of the generalization; to study the
consistency of generalization estimators for different classes of
systems and identify other useful components for instantiating the
bootstrap-based approach for estimating the generalization, including
log sampling methods and generalization measures over known systems;
and to explore the quality of different bootstrap-based estimators of
the generalization to overcome problems associated with noisy logs.
\medskip

\noindent
\textbf{Acknowledgment.}
Artem Polyvyanyy was in part supported by the Australian Research Council project DP180102839.
\updated{A presentation of this work from an earlier stage of the research project is available at \url{https://youtu.be/8I-87iGCzNI}.}



\vspace{-2mm}
\bibliography{bibliography}


\end{document}